\documentclass[remotesensing,article,accept,pdftex,moreauthors]{Definitions/mdpi}
\usepackage{subcaption}
\usepackage{amsmath}
\usepackage{multirow}
\usepackage{float}
\usepackage{wrapfig}
\usepackage{changes}
\usepackage[labelformat=simple]{subcaption}

\DeclareCaptionLabelFormat{subcaptionlabel}{\normalfont(\textbf{#2}\normalfont)}
\firstpage{1}
\pubvolume{14}
\issuenum{9}
\articlenumber{2254}
\pubyear{2022}
\copyrightyear{2022}
\externaleditor{Academic Editors: Mohammad Awrangjeb, Jiaojiao Tian, Qin Yan, Beril Sirmacek and Nusret Demir} 
\datereceived{24 March 2022}
\dateaccepted{2 May 2022}
\datepublished{7 May 2022}
\hreflink{https://\linebreak doi.org/10.3390/rs14092254} 

\Title{{City3D:} Large-Scale Building Reconstruction from Airborne LiDAR Point Clouds}

\TitleCitation{City3D: Large-Scale Building Reconstruction from Airborne LiDAR Point Clouds}

\Author{Jin Huang~\href{https://orcid.org/0000-0002-9804-0195}{\orcidicon}, {Jantien Stoter~\href{https://orcid.org/0000-0002-1393-7279}{\orcidicon}, Ravi Peters} and Liangliang Nan *\href{https://orcid.org/0000-0002-5629-9975}{\orcidicon}}

\AuthorNames{Jin Huang, Jantien Stoter, Ravi Peters and Liangliang Nan}

\AuthorCitation{Huang, J.; Stoter, J.; Peters, R.; Nan, L.}

\address[1]{%
3D Geoinformation Research Group, Faculty of Architecture and the Built Environment, Delft University \mbox{of Technology}, 2628 BL Delft, The Netherlands; j.huang-1@tudelft.nl (J.H.);  	j.e.stoter@tudelft.nl (J.S.); r.y.peters@tudelft.nl (R.P.)}

\corres{\hangafter=1 \hangindent=1.05em \hspace{-0.82em}Correspondence: liangliang.nan@tudelft.nl}

\abstract{We present a fully automatic approach for reconstructing compact 3D building models from large-scale airborne point clouds.
A major challenge of urban reconstruction from airborne LiDAR point clouds lies in that the vertical walls are typically missing.
Based on the observation that urban buildings typically consist of planar roofs connected with vertical walls to the ground, we propose an approach to infer the vertical walls directly from the data.
With the planar segments of both roofs and walls, we hypothesize the faces of the building surface, and the final model is obtained by using an extended hypothesis-and-selection-based polygonal surface reconstruction framework. Specifically, we introduce a new energy term to encourage roof preferences and two additional hard constraints into the optimization step to ensure correct topology and enhance detail recovery.
Experiments on various large-scale airborne LiDAR point clouds have demonstrated that the method is superior to the state-of-the-art methods in terms of reconstruction accuracy and robustness.
In addition, we have generated a new dataset with our method consisting of the point clouds and 3D models of 20k real-world buildings. We believe this dataset can stimulate research in urban reconstruction from airborne LiDAR point clouds and the use of 3D city models in urban applications.
} 


\begin{document}


\section{Introduction}
\label{sec:introduction}
Digitizing urban scenes is an important research problem in computer vision, computer graphics, and photogrammetry communities.
Three-dimensional models of urban buildings have become the infrastructure for a variety of real-world applications such as visualization~\citep{yao20183dcitydb}, simulation~\citep{zhivov2017planning,stoter2020automated,widl2021linking}, navigation~\citep{cappelle2012virtual}, and entertainment~\citep{kargas2019using}.
These applications typically require high-accuracy and compact 3D building models of large-scale \mbox{urban environments}.

Existing urban building reconstruction methods strive to bring in a great level of detail and automate the process for large-scale urban environments. Interactive reconstruction techniques are successful in reconstructing accurate 3D building models with great detail~\cite{nan2010smartboxes,nan2015template}, but they require either high-quality laser scans as input or considerable amounts of user interaction. These methods can thus hardly be applied to large-scale urban scenes.
To facilitate practical applications that require large-scale 3D building models, researchers have attempted to address the reconstruction challenge using various data sources~\cite{zhou20123d,haala2015extracting,verdie2015lod,li2016reconstructing,buyukdemircioglu2018semi,bauchet2019,li2019modelling,ledoux20213dfier}.
Existing methods based on aerial images~\cite{haala2015extracting,li2016reconstructing,buyukdemircioglu2018semi} and dense triangle meshes~\cite{verdie2015lod} typically require good coverage of the buildings, which imposes challenges in data acquisition~\cite{zhou2020survey}.
Approaches based on airborne LiDAR point clouds alleviate data acquisition issues. However, the accuracy and geometric details are usually compromised~\cite{zhou20123d,bauchet2019,li2019modelling,ledoux20213dfier}.
Following previous works using widely available airborne LiDAR point clouds, we strive to recover desired geometric details of real-world buildings while ensuring topological correctness, reconstruction accuracy, and good efficiency.

The challenges for large-scale urban reconstruction from airborne LiDAR point \linebreak clouds include:

\begin{itemize}
\item {Building instance segmentation}. Urban scenes are populated with diverse objects, such as buildings, trees, city furniture, and dynamic objects (e.g., vehicles and pedestrians). The cluttered nature of urban scenes poses a severe challenge to the identification and separation of individual buildings from the massive point clouds. This has drawn considerable attention in recent years~\cite{qi2017pointnet,thomas2019kpconv}.

\item {Incomplete data}. Some important structures (e.g., vertical walls) of buildings are typically not captured in airborne LiDAR point clouds due to the restricted positioning and moving trajectories of airborne scanners.

\item {Complex structures}. Real-world buildings demonstrate complex structures with varying styles. However, limited cues about structure can be extracted from the sparse and noisy point clouds, which further introduces ambiguities in obtaining topologically correct surface models.

\end{itemize}

In this work, we address the above challenges with the following strategies.
Firstly, we address the building instance segmentation challenge by separating individual buildings using increasingly-available vectorized building footprint data.
Secondly, we exploit prior knowledge about the structures of buildings to infer their vertical planes. Based on the fact that vertical planes in airborne LiDAR point clouds are typically walls connecting the piecewise planar roofs to the ground, we propose an algorithm to infer the vertical planes from incomplete point clouds.
Our method has the option to extrude outer walls directly from the given building footprint.
Finally, we approach surface reconstruction by introducing the inferred vertical planes as constraints into an existing hypothesis-and-selection-based polygonal surface reconstruction framework~\cite{nan2017polyfit}, which favors good fitting to the input point cloud, encourages compactness, and enforces manifoldness of the final model (see Figure~\ref{fig:city_result} for an example of the reconstruction results). The main contributions of this work include:

\begin{itemize}
\item A robust framework for fully automatic reconstruction of large-scale urban buildings from airborne LiDAR point clouds.
\item An extension of an existing hypothesis-and-selection-based surface reconstruction method for buildings, which is achieved by introducing a new energy term to encourage roof preferences and two additional hard constraints to ensure correct topology and enhance detail recovery.
\item A novel approach for inferring vertical planes of buildings from airborne LiDAR point clouds, for which we introduce an optimal-transport method to extract polylines from 2D bounding contours.
\item A new dataset consisting of the point clouds and reconstructed surface models of 20 k real-world buildings.

\end{itemize}

\vspace{-6pt}
\begin{figure}[H]

\begin{subfigure}[t]{0.9\linewidth}
\includegraphics[width=\linewidth]{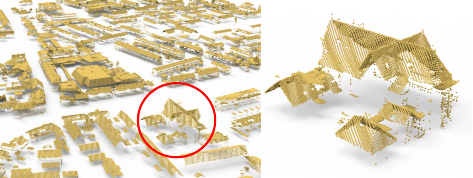}
\caption{\centering Input airborne LiDAR point cloud.}
\end{subfigure}
\caption{\emph{Cont}.}
\label{fig:city_result}
\end{figure}
\begin{figure}[H]\ContinuedFloat
\begin{subfigure}[t]{0.9\linewidth}
\includegraphics[width=\linewidth]{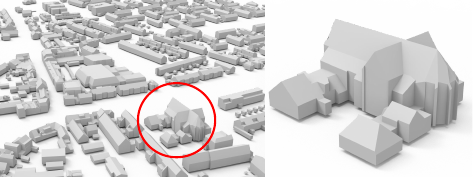}
\caption{\centering Our reconstruction result}
\end{subfigure}
\caption{The automatic reconstruction result of all the buildings in a large scene from the AHN3 {dataset}~\cite{AHN3_2018}.} 
\label{fig:city_result}
\end{figure}

\section{Related Work}\label{sec:relatedwork}

A large volume of methods for urban building reconstruction has been proposed. In this section, we mainly review the techniques relevant to the key components of our method. Since our method relies on footprint data for extracting building instances from the massive point clouds of large scenes, and it can also be used for footprint extraction, we also discuss related techniques in footprint extraction.

\textbf{{Roof primitive extraction}}. 
The commonly used method for extracting basic primitives (e.g., planes and cylinders) from point clouds is random sample consensus (RANSAC)~\citep{fischler1981random} and its variants ~\citep{schnabel2007efficient,zuliani2005multiransac}, which are robust against noise and outliers.
Another group of widely used methods is based on region growing~\cite{rabbani2006segmentation,sun2013aerial,chen2017topologically}, which assumes roofs are piece-wise planar and iteratively propagates planar regions by advancing the boundaries.
The main difference between existing region growing methods lies in the generation of seed points and the criteria for region expansion.
In this paper, we utilize an existing region growing method to extract roof primitives given its simplicity and robustness, which is detailed in Rabbani et al.~\citep{rabbani2006segmentation}.

\textbf{{Footprint extraction}}.
Footprints are 2D outlines of buildings, capturing the geometry of outer walls projected onto the ground plane.
Methods for footprint extraction commonly project the points to a 2D grid and analyze their distributions~\citep{meng2009morphology}.
Chen et al.~\citep{chen2017topologically} detect rooftop boundaries and cluster them by taking into account topological consistency between the contours.
To obtain simplified footprints, polyline simplification methods such as the Douglas-Peucker algorithm~\citep{douglas1973algorithms} are commonly used to reduce the complexity of the extracted contours~\citep{zhang2006automatic,li2016reconstructing,xiong2016footprint}.
To favor structural regularities, Zhou and Neumann~\citep{zhou2008fast} compute the principal directions of a building and regularize the roof boundary polylines along with these directions.
Following these works, we infer the vertical planes of a building by detecting its contours from a heightmap generated from a 2D projection of the input points. The contour polylines are then regularized by orientation-based clustering followed by an adjustment step.

\textbf{{Building surface reconstruction}}.
This type of methods aims at obtaining a simplified surface representation of buildings by exploiting geometric cues, e.g., planar primitives and their boundaries~\citep{dorninger2008comprehensive,zhou2008fast,lafarge2012creating,xiao2015building,yi2017urban,li2019modelling}.
Zhou and Neumann~\citep{zhou20102} approached this by simplifying the 2.5D TIN (triangulated irregular network) of buildings, which may result in artifacts in building contours due to its limited capability in capturing complex topology. To address this issue, the authors proposed an extended 2.5D contouring method with improved topology control~\citep{zhou20112}.
To cope with missing walls,~\citet{chauve2010robust} also incorporated additional primitives inferred from the point clouds.
Another group of building surface reconstruction methods involves predefined building parts, commonly known as model-driven approaches~\citep{lafarge2008structural,xiong2014graph}.
These methods rely on templates of known roof structures and deform-to-fit the templates to the input points. Therefore, the results are usually limited to the predefined shape templates, regardless of the diverse and complex nature of roof structures or high intraclass variations.
Given the fact that buildings demonstrate mainly piecewise planar regions, methods have also been proposed to obtain an arrangement of extracted planar primitives to represent the building geometry~\citep{li2016boxfitting,nan2017polyfit,bauchet2020kinetic,fang2020connect}.
These methods first detect a set of planar primitives from the input point clouds and then hypothesize a set of polyhedral cells or polygonal faces using the supporting planes of the extracted planar primitives.
Finally, a compact polygonal mesh is extracted from the hypothesized cells or faces.
These methods focus on the assembly of planar primitives, for which obtaining a complete set of planar primitives from airborne LiDAR point clouds is still a challenge.

In this work, we extend an existing hypothesis-and-selection-based general polygonal surface reconstruction method~\citep{nan2017polyfit} to reconstruct buildings that consist of piecewise planar roofs connected to the ground by vertical walls.
We approach this by introducing a novel energy term and a few hard constraints specially designed for buildings to ensure correct topology and decent details.

\section{Methodology}\label{sec:method}

\subsection{Overview}
\label{sec:overview}

\begin{figure}[H]

\includegraphics[width=13.5cm]{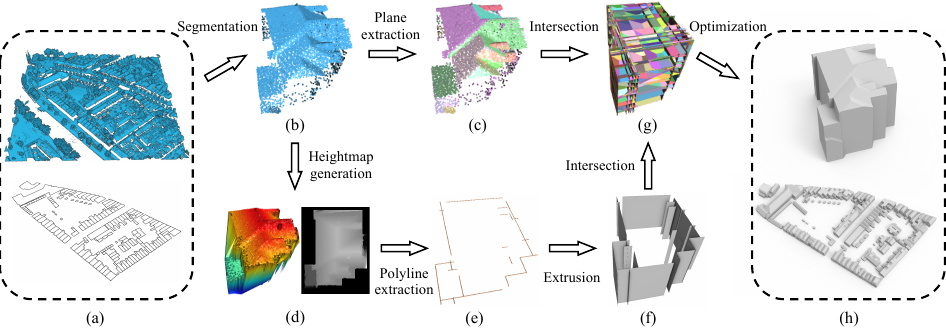}
\caption{The pipeline of the proposed method (only one building is selected to illustrate the workflow). (\textbf{a}) Input point cloud and corresponding footprint data. (\textbf{b}) A building extracted from the input point cloud using its footprint polygon. (\textbf{c}) Planar segments extracted from the point cloud. (\textbf{d}) The heightmap (right) generated from the TIN (left, colored as a height field). (\textbf{e}) The polylines extracted from the heightmap. (\textbf{f}) The vertical planes obtained by extruding the inferred polylines. (\textbf{g}) The hypothesized building faces generated using both the extracted planes and inferred vertical planes. (\textbf{h}) The final model obtained through optimization.
}
\label{fig:pipeline}
\end{figure}

The proposed approach takes as input a raw airborne LiDAR point cloud of a large urban scene and the corresponding building footprints, and it outputs 2-manifold and watertight 3D polygonal models of the buildings in the scene.
Figure~\ref{fig:pipeline} shows the pipeline of the proposed method.
It first extracts the point clouds of individual buildings by projecting all points onto the ground plane and collecting the points lying inside the footprint polygon of each building.
Then, we reconstruct a compact polygonal model from the point cloud of each building.

\vspace{-6pt}

Our reconstruction of a single building is based on the hypothesis-and-selection-based framework of PolyFit~\cite{nan2017polyfit}, which is for reconstructing general piecewise-planar objects from a set of planar segments extracted from the point cloud.
Our method exploits not only the planar segments directly extracted from the point cloud but also the vertical planes inferred from the point cloud.
From these two types of planar primitives, we hypothesize the faces of the building.
The final model is then obtained by choosing the optimal subset of the faces through optimization.

The differences between our method and PolyFit are: (1) our method is dedicated to reconstructing urban buildings, and it makes use of vertical planes as hard constraints, for which we propose a novel algorithm for inferring the vertical planes of buildings that are commonly missing in airborne LiDAR point clouds. (2) We introduce a new \textit{{roof preference}} 
energy term and two additional hard constraints into the optimization to ensure correct topology and enhance detail recovery. In the following sections, we detail the key steps of our method with an emphasis on the processes that differ from PolyFit~\cite{nan2017polyfit}.

\subsection{Inferring Vertical Planes}
\label{subsec:InferWalls}

With airborne LiDAR point clouds, important structures like vertical walls of a building are commonly missed due to the restricted positioning and moving trajectories of the scanner.
In contrast, the roof surfaces are usually well captured.
This inspired us to infer the missing walls from the available points containing the roof surfaces.
We infer the vertical planes representing not only the outer walls but also the vertical walls within the footprint of a building. We achieve this by generating a 2D rasterized height map from its 3D points and looking for the contours that demonstrate considerable variations in the height values. To this end, an optimal-transport method is proposed to extract closed polylines from the contours. The polylines are then extruded to obtain the vertical walls.
The process for inferring the vertical planes is outlined in Figure~\ref{fig:pipeline}{d--f}.

Specifically, after obtaining the point cloud of a building, we project the points onto the ground plane, from which we create a heigh{t} map.
To cope with the non-uniform distribution of the points (e.g., some regions have holes while others may have repeating points), we construct a Triangulated Irregular Network (TIN) model using 2D Delaunay triangulation. The TIN model is a continuous surface and naturally completes the missing regions.
Then, a height map is generated by rasterizing the TIN model with a specified resolution $r$.
The issue of small holes in the height maps (due to uneven distribution of roof points) is further alleviated by image morphological operators while preserving the shape and size of the building~\citep{huang2013generative}.
After that, a set of contours are extracted from the height map using the Canny detector~\citep{canny1986computational}, which serves as the initial estimation of the vertical planes. We propose an optimal-transport method to extract polylines from the initial set \mbox{of contours}.

\textbf{{Optimal-transport method for polyline extraction}}.
The initial set of contours are discrete pixels, denoted as $S$, from which we would like to extract simplified polylines that best describe the 2D geometry of $S$.
Our optimal-transport method for extracting polylines from $S$ works as follows.
First, a 2D Delaunay triangulation $T_0$ is constructed from the discrete points in $S$.
Then, the initial triangulation $T_0$ is simplified through iterative edge collapse and vertex removal operations. In each iteration, the most suitable vertex to be removed is determined in a way such that the following conditions are met:
\begin{itemize}
\item The maximum Hausdorff distance from the simplified mesh $T_0$ to $S$ is less than a distance threshold $\epsilon_d$.
\item The increase of the total transport cost~\citep{de2011optimal} between $S$ and $T_0$ is kept at a minimum.
\end{itemize}

In each iteration, a vertex satisfying the above conditions is removed from $T_0$ by edge collapse, and the overall transportation cost is updated.

As the iterative simplification process continues, the overall transportation cost will increase.
The edge collapse operation stops until no vertex can be further removed, or the overall transportation cost has increased above a user-specified tolerance $\epsilon_c$.
After that, we apply an edge filtering step~\citep{de2011optimal} to eliminate small groups of undesirable edges caused by noise and outliers.
Finally, the polylines are derived from the remaining vertices and edges of the simplified triangulation using the procedure described in~\cite{de2011optimal}.
Compared to~\cite{de2011optimal}, our method not only minimizes the total transport cost but also provides control over local geometry, ensuring that the distance between every vertex in the final polylines and the initial contours is smaller than the specified distance threshold $\epsilon_d$.

\begin{figure}[H]
\captionsetup[subfigure]{justification=centering}
\centering
\begin{subfigure}[t]{0.45\linewidth}
\includegraphics[width=\linewidth]{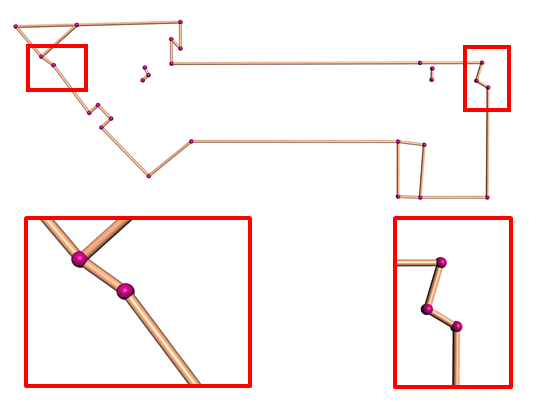}
\caption{Before (28 segments)}
\end{subfigure}
\hspace{0.6cm}
\begin{subfigure}[t]{0.45\linewidth}
\includegraphics[width=\linewidth]{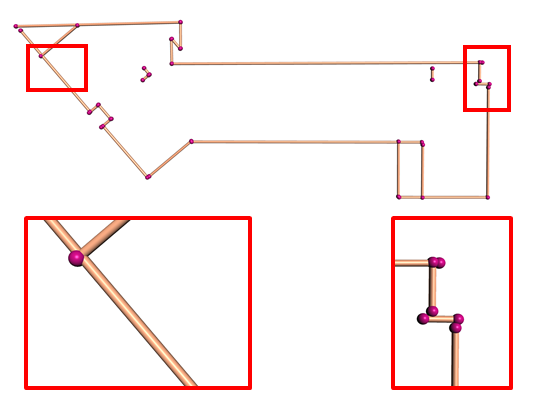}
\caption{After (22 segments)}
\end{subfigure}
\caption{The effect of the clustering-based regularity enhancement on the polylines inferring the vertical walls. (\textbf{a}) Before regularity enhancement. (\textbf{b}) After regularity enhancement.}
\label{fig:regularity_enhance}
\end{figure}

\textbf{{Regularity enhancement}}.
Due to noise and uneven point density in the point cloud, the polylines generated by the optimal-transport algorithm are unavoidably inaccurate and irregular (see Figure~\ref{fig:regularity_enhance}a), which often leads to artifacts in the final reconstruction.
We alleviate these artifacts by enforcing structure regularities that commonly dominate urban buildings.
{We consider the structure regularities, namely parallelism, collinearity, and orthogonality, defined by~\cite{li2021relation}. Please note that since all the lines will be extruded vertically to obtain the vertical planes, the verticality regularity will inherently be satisfied.}
We propose a clustering-based method to identify the groups of line segments that potentially satisfy these regularities. Our method achieves structure regularization in two steps: clustering and adjustment.

\vspace{-6pt}
\textit{{Clustering}}. In this work, we cluster the line segments of the polylines generated by the optimal-transport algorithm based on their orientation and pairwise Euclidean distance~\cite{schubert2017dbscan}. The pairwise Euclidean distance is measured by the minimum distance between a line segment and the supporting line of the other line segment.

\textit{{Adjustment}}. For each cluster that contains multiple line segments, we compute its average direction. Then each line segment in the cluster is adjusted to align with the average direction.
In case the building footprint is provided, the structure regularity can be further improved by aligning the segments with the edges in the footprint.
After average adjustment, the near-collinear and near-orthogonal line segments are adjusted to be perfectly collinear and orthogonal, respectively (we use an angle threshold of 20\textdegree).

After regularity enhancement, the vertical planes of the building can be obtained by vertical extrusion of the regularized polylines. The effect of the regularity enhancement is demonstrated in Figure~\ref{fig:regularity_enhance}, from which we can see that it significantly improves structure regularity and reduces the complexity of the building outlines.

\subsection{Reconstruction}
\label{sec:reconstrtuction}

Our surface reconstruction involves two types of planar primitives, i.e., vertical planes inferred in the previous step (see Section~\ref{subsec:InferWalls}) and roof planes directly extracted from the point cloud.
Unlike PolyFit~\cite{nan2017polyfit} that hypothesizes faces by computing pairwise intersections using all planar primitives, we compute pairwise intersections using only the roof planes, and then the resulted faces are cropped with the outer vertical planes (see Figure~\ref{fig:pipeline}g). This process ensures that the roof boundaries of the reconstructed building can be precisely connected with the inferred vertical walls. Additionally, since the object to be reconstructed is a real-world building, we introduce a \textit{{roof preference}} energy term and a set of new hard constraints specially designed for buildings into the original formulation. Specifically, our objective for obtaining the model faces $F^*$ can be written as
\begin{equation}
\label{eq:optimization}
F^*= \arg \min_{X} \lambda_{d}E_{d}+\lambda_{c} E_{c}+\lambda_{r} E_{r},
\end{equation}
where $X = \{x_i | x_i \in \{0, 1\}\}$ denotes the binary variables for the faces (1 for \textit{{selected}} and 0 otherwise). $E_d$ is the data fitting term that encourages selecting faces supported by more points, and $E_{c}$ is the model complexity term that favors simple planar structures. For more details about the data fitting term and the model complexity term, please refer to the original paper of PolyFit~\cite{nan2017polyfit}. In the following part, we elaborate on the new energy term and hard constraints.

\textbf{{New energy term}}: \textit{{roof preference}}. We have observed in rare cases that a building in aerial point clouds may demonstrate more than one layer of roofs, e.g., semi-transparent or overhung roofs. In such a case, we assume a higher roof face is always preferable to the ones underneath. We formulate this preference as an additional energy term called \textit{{roof preference}}, which is defined as
\begin{equation}\label{term3}
E_{r}=\frac{1}{\left| F\right| }\sum_{i=1}^{|F|}x_{i} \cdot \frac{z_{max}-z_{i}}{z_{max}-z_{min}}
\end{equation}
where $z_{i} $ denotes the $Z$ coordinate of the centroid of a hypothsized face $f_{i}$. $z_{max}$ and $z_{min}$ are, respectively, the highest and lowest $Z$ coordinates of the building points. $|F|$ denotes the total number of hypothesized faces.

\textbf{{New hard constraints}}.
We impose two hard constraints to enhance the topological correctness of the final reconstruction.

\begin{figure}[H]

\begin{subfigure}[t]{5cm}
\includegraphics[width=\linewidth]{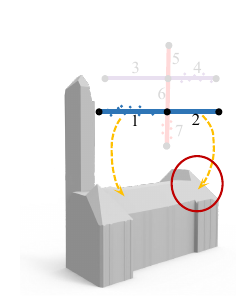}
\caption{\centering}
\end{subfigure}
\hspace{0.6cm}
\begin{subfigure}[t]{5cm}
\includegraphics[width=\linewidth]{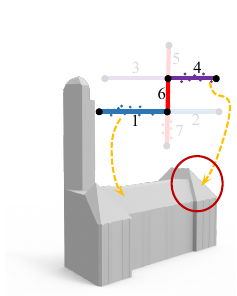}
\caption{\centering}
\end{subfigure}
\caption{The effect of the \textit{{face prior}} constraint. The insets illustrate the assembly of the hypothesized faces in the corresponding marked regions (each line segment denotes a hypothesized face, and line segments of the same color represent faces derived from the same planar primitive).
(\textbf{a}) Reconstruction without the \textit{{face prior}} constraint. (\textbf{b}) Reconstruction with the \textit{{face prior}} constraint, for which faces 1 and 4 both satisfy the \textit{{face prior}} constraint. The numbers 1--7 denote the 7 candidate faces.
}
\label{sub:face_prior_constraint}
\end{figure}

\begin{itemize}
\item  \textit{{Single-layer roof}}. This constraint ensures that the reconstructed 3D model of a real-world building has a single layer of roofs, which can be written as,
\begin{align*}
\sum\limits_{k \in V(f_{i})} x_{k}=1 , & (1 \leq i \leq |F|)
\end{align*}
where $V(f_{i})$ denotes the set of hypothesized faces that have overlap with face $f_i \in F$ in the vertical direction.

\item \textit{{Face prior}}. This constraint enforces that for all the derived faces from the same planar segment, the one with the highest confidence value is always selected as a prior. Here, the confidence of a face is measured by the number of its supporting points. This constraint can be simply written as
\begin{align*}
x_{l}=1,
\end{align*}
where $x_{l}$ is the variable whose value denotes the status of the most confident face $f_l$ of a planar segment.
This constraint resolves ambiguities if two hypothesized faces are near coplanar and close to each other, which preserves finer geometric details.
The effect of this constraint is demonstrated in Figure~\ref{sub:face_prior_constraint}.

\end{itemize}
\vspace{-15pt}

The final surface model of the building can be obtained by solving the optimization problem given in Equation~(\ref{eq:optimization}), subject to the \textit{{single-layer roof}} and \textit{{face prior}} hard constraints.

\section{Results and Evaluation}
\label{sec:results}

Our method is implemented in C++ using CGAL~\citep{cgal:eb-20a}.
All experiments were conducted on a desktop PC with a 3.5 GHz AMD Ryzen Threadripper 1920X and 64 GB RAM.

\subsection{Test Datasets}
We have tested our method on three datasets of large-scale urban point clouds including more than 20~k buildings.
\begin{itemize}
\item AHN3~\citep{AHN3_2018}. An {openly available }country-wide airborne LiDAR point cloud dataset covering the entire Netherlands, with an average point density of 8 points/m$^2$. The corresponding footprints of the buildings are obtained from the Register of Buildings and Addresses (BAG)~\citep{BAG_2018}. {The geometry of footprint is acquired from aerial photos and terrestrial measurements with an accuracy of 0.3 m}.
{The polygons in the BAG represent the outlines of buildings as their outer walls seen from above, which are slightly different from footprints. We still use `footprint' in this paper. }

\item DALES~\citep{varney2020dales}. A large-scale aerial point cloud dataset consisting of forty scenes spanning an area of 10 km$^2$, with instance labels of 6 k buildings.
{The data was collected using a Riegl Q1560 dual-channel system with a flight altitude of 1300  m  above ground and a speed of 72 m/s.
Each area was collected by a minimum of 5 laser pulses per meter in four directions.
The LiDAR swaths were calibrated using the BayesStripAlign 2.0 software and registered, taking both relative and absolute errors into account and correcting for altitude and positional errors. }
The average point density is 50~points/m$^2$.
No footprint data is available in this dataset.

\item Vaihingen~\cite{rottensteiner2012isprs}. An airborne LiDAR point cloud dataset published by ISPRS, which has been widely used in semantic segmentation and reconstruction of urban scenes.
{The data were obtained using a Leica ALS50 system with 45\textdegree~field of view and a mean flying height above ground of 500 m.
The average strip overlap is 30\% and multiple pulses were recorded.
The point cloud was pre-processed to compensate for systematic offsets between the strips.}
We use in our experiments a training set that contains footprint information and covers an area of 399 m $\times$ 421 m with 753 k points. The average point density is 4 points/m$^2$.

\end{itemize}

\subsection{Reconstruction Results}
\textbf{{Visual results}}.
We have used our method to reconstruct more than 20~k buildings from the aforementioned three datasets.
For the AHN3~\citep{AHN3_2018} and Vaihingen~\cite{rottensteiner2012isprs} datasets, the provided footprints were used for both building instance segmentation and extrusion of the outer walls. Our inferred vertical planes were used to complete the missed inner walls.
For the DALES~\citep{varney2020dales} dataset, we used the provided instance labels to extract building instances, and we used our inferred vertical walls for the reconstruction.

\begin{figure}[H]

\includegraphics[width=13cm]{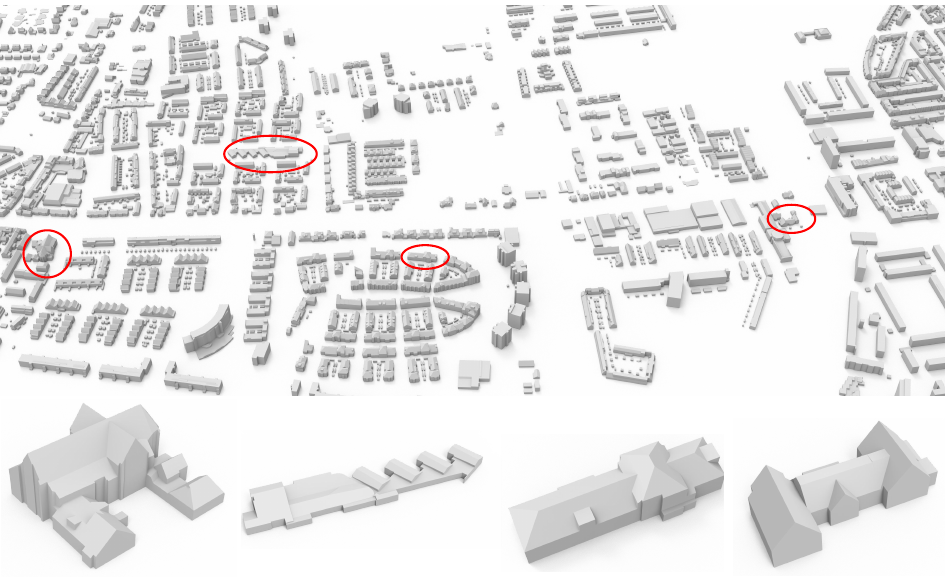}
\caption{Reconstruction of a large scene from the AHN3 {dataset}~\cite{AHN3_2018}.}
\label{fig:city_area} 
\end{figure}

\begin{figure}[H]

\includegraphics[width=13cm]{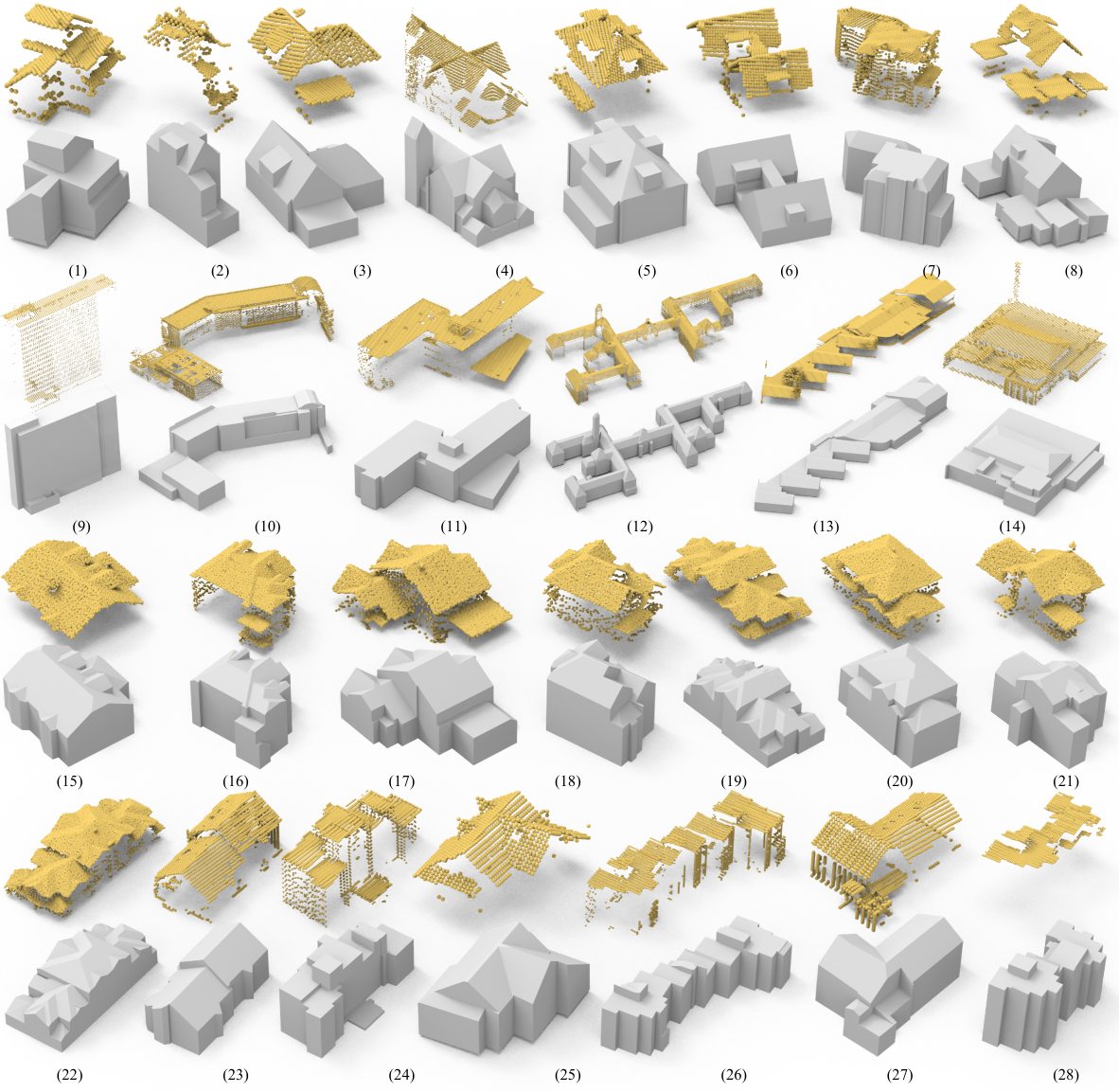}
\caption{The reconstruction results of a set of buildings from various dataset. (\textbf{1}--\textbf{14}) are from the AHN3 {dataset}~\cite{AHN3_2018}, (\textbf{15}--\textbf{22}) are from the DALES {dataset}~\citep{varney2020dales}, (\textbf{23}--\textbf{28}) are from the Vaihingen {dataset}~\cite{rottensteiner2012isprs}.}
\label{fig:individual_buildings_results}
\end{figure}

Figures~\ref{fig:city_result} and~\ref{fig:city_area} show the 3D reconstruction of all buildings in two large scenes from the AHN3 dataset~\cite{AHN3_2018}. For the buildings reconstructed in Figure~\ref{fig:city_result}, their models are simplified polygonal meshes with an average face count of 34.
To better reveal the quality of our reconstructed building models, we demonstrate in Figure~\ref{fig:individual_buildings_results} a set of individual buildings reconstructed from the three test datasets.
From these visual results, we can see that although the buildings have diverse structures of different styles, and the input point clouds have varying densities and different levels of noise, outliers, and missing data, our method succeeded in obtaining visually plausible reconstruction results.
These experiments also indicate that our approach is successful in inferring the vertical planes of buildings from airborne LiDAR point clouds and it is effective to include these planes in the 3D reconstruction of urban buildings.

\vspace{-9pt}

\begin{figure}[H]

\includegraphics[width=0.8\linewidth]{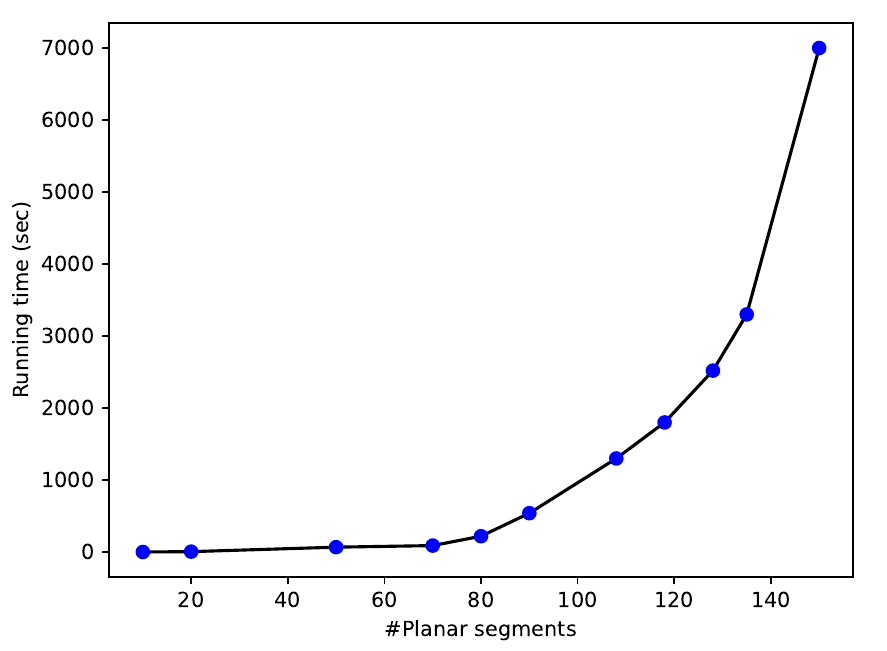}
\caption{ {The running time of our method with respect to the number of the detected planar segments. These statistics are obtained by testing on the AHN3 dataset.}}
\label{fig:timing}
\end{figure}

\textbf{{Quantitative results}}.
We have also evaluated the reconstruction results quantitatively.
Since ground-truth reconstruction is not available for all buildings in the three datasets, we chose to use the commonly used accuracy measure, Root Mean Square Error (RMSE), to quantify the quality of each reconstructed model.
In the context of surface reconstruction, RMSE is defined as the square root of the average of squared Euclidean distances from the points to the reconstructed model.
In Table~\ref{tab:statistics}, we report the statistics of our quantitative results on the buildings shown in Figure~\ref{fig:individual_buildings_results}.
We can see that our method has obtained good reconstruction accuracy, i.e., the RMSE for all buildings is between 0.04 m to 0.26 m, which is quite promising for 3D reconstruction of real-world buildings from noisy and sparse airborne LiDAR point clouds.
As observed from the number of faces column of Table~\ref{tab:statistics}, our results are simplified polygonal models and are more compact than those obtained from commonly used approaches such as the Poisson surface reconstruction method~\cite{kazhdan2006poisson} (that produces dense triangles).
Table~\ref{tab:statistics} also shows that the running times for most buildings are less than 30 s.
The reconstruction of the large complex building shown in Figure~\ref{fig:individual_buildings_results} (12) took 42 min.
{This long reconstruction time is due to that our method computes the pairwise intersection of the detected planar primitives and inferred vertical planes, and it generates a large number of candidate faces and results in a large optimization problem~\cite{nan2017polyfit} (see also Section~\ref{sec:limitation}). The running time with respect to the number of detected planar segments for the reconstruction of more buildings is reported in Figure~\ref{fig:timing}.}

\begin{table}[H]
\caption{Statistics on the reconstructed buildings shown in Figure~\ref{fig:individual_buildings_results}. For each building, the number of points in the input, number of faces in the reconstructed model, fitting error (i.e., RMSE in meters), and running time (in seconds) are reported.}
\label{tab:statistics}
\setlength{\cellWidtha}{\textwidth/6-2\tabcolsep-0in}
\setlength{\cellWidthb}{\textwidth/6-2\tabcolsep-0in}
\setlength{\cellWidthc}{\textwidth/6-2\tabcolsep-0in}
\setlength{\cellWidthd}{\textwidth/6-2\tabcolsep-0in}
\setlength{\cellWidthe}{\textwidth/6-2\tabcolsep-0in}
\setlength{\cellWidthf}{\textwidth/6-2\tabcolsep-0in}
\scalebox{1}[1]{\begin{tabularx}{\textwidth}{>{\centering\arraybackslash}m{\cellWidtha}>{\centering\arraybackslash}m{\cellWidthb}>{\centering\arraybackslash}m{\cellWidthc}>{\centering\arraybackslash}m{\cellWidthd}>{\centering\arraybackslash}m{\cellWidthe}>{\centering\arraybackslash}m{\cellWidthf}}
\toprule
{\textbf{Dataset}} &
{\textbf{Model}} &
\textbf{\#Points} &
\textbf{\#Faces} &
\textbf{RMSE}\linebreak\textbf{(m)} & \textbf{Time}\linebreak\textbf{(s)} \\ \midrule
\multirow{18.5}{*}{AHN3}     & (1)  & 732    & 23   & 0.07 & 3    \\ \cmidrule{2-6}
& (2)  & 532    & 42   & 0.12 & 4    \\ \cmidrule{2-6}
& (3)  & 1165   & 31   & 0.04 & 3    \\ \cmidrule{2-6}
& (4)  & {20,365}  
& 127  & 0.15 & 62   \\ \cmidrule{2-6}
& (5)  & 1371   & 48   & 0.04 & 5    \\ \cmidrule{2-6}
& (6)  & 1611   & 45   & 0.06 & 4    \\ \cmidrule{2-6}
& (7) & 3636   & 68   & 0.21 & 18   \\ \cmidrule{2-6}
& (8) & 2545   & 52   & 0.04 & 8    \\ \cmidrule{2-6}
& (9) & {15,022}  & 63   & 0.11 & 28   \\ \cmidrule{2-6}
& (10)  & {23,654}  & 262  & 0.26 & 115  \\ \cmidrule{2-6}
& (11) & {13,269}  & 102  & 0.11 & 34   \\ \cmidrule{2-6}
& (12)  & {155,360} & 1520 & 0.09 & 2520 \\ \cmidrule{2-6}
& (13) & {24,027}  & 176  & 0.24 & 141  \\  \cmidrule{2-6}
& (14)  & {28,522}  & 227  & 0.15 & 78   \\ \midrule
\multirow{11}{*}{DALES}     & (15) & 8662   & 39   & 0.04 & 11   \\ \cmidrule{2-6}
& (16) & {11,830}  & 73   & 0.1  & 8    \\ \cmidrule{2-6}
& (17) & {10,673}  & 47   & 0.07 & 7    \\ \cmidrule{2-6}
& (18) & 7594   & 33   & 0.07 & 14   \\ \cmidrule{2-6}
& (19) & {13,060}  & 278  & 0.05 & 145  \\ \cmidrule{2-6}
& (20) & {11,114}  & 55   & 0.06 & 24   \\ \cmidrule{2-6}
& (21) & 8589   & 51   & 0.06 & 15   \\ \cmidrule{2-6}
& (22) & {18,909}  & 282  & 0.08 & 86   \\ \bottomrule
\end{tabularx}}

\end{table}

\begin{table}[H]\ContinuedFloat
\caption{\emph{Cont}.}
\label{tab:statistics}
\setlength{\cellWidtha}{\textwidth/6-2\tabcolsep-0in}
\setlength{\cellWidthb}{\textwidth/6-2\tabcolsep-0in}
\setlength{\cellWidthc}{\textwidth/6-2\tabcolsep-0in}
\setlength{\cellWidthd}{\textwidth/6-2\tabcolsep-0in}
\setlength{\cellWidthe}{\textwidth/6-2\tabcolsep-0in}
\setlength{\cellWidthf}{\textwidth/6-2\tabcolsep-0in}
\scalebox{1}[1]{\begin{tabularx}{\textwidth}{>{\centering\arraybackslash}m{\cellWidtha}>{\centering\arraybackslash}m{\cellWidthb}>{\centering\arraybackslash}m{\cellWidthc}>{\centering\arraybackslash}m{\cellWidthd}>{\centering\arraybackslash}m{\cellWidthe}>{\centering\arraybackslash}m{\cellWidthf}}
\toprule
{\textbf{Dataset}} &
{\textbf{Model}} &
\textbf{\#Points} &
\textbf{\#Faces} &
\textbf{RMSE}\linebreak\textbf{(m)} & \textbf{Time}\linebreak\textbf{(s)} \\ \midrule
\multirow{8}{*}{Vaihingen}
& (23) & 7701   & 51   & 0.24 & 25   \\ \cmidrule{2-6}
& (24) & 6845   & 99   & 0.12 & 8    \\ \cmidrule{2-6}
& (25) & 1007   & 24   & 0.11 & 2    \\ \cmidrule{2-6}
& (26) & {11,591}  & 206  & 0.17 & 10   \\ \cmidrule{2-6}
& (27) & 4026   & 42   & 0.26 & 6    \\ \cmidrule{2-6}
& (28) & 5059   & 61   & 0.22 & 9    \\
\bottomrule
\end{tabularx}}

\end{table}
\vspace{-12pt}
\textbf{{New dataset}}.
Our method has been applied to city-scale building reconstruction. The results are released as a new dataset consisting of 20~k buildings (including the reconstructed 3D models and the corresponding airborne LiDAR point clouds).
We believe this dataset can stimulate research in urban reconstruction from airborne LiDAR point clouds and the use of 3D city models in urban applications.

\subsection{Parameters}
Our method involves a few parameters that are empirically set to fixed values for all experiments, i.e., the distance threshold $\epsilon_d = 0.25$ and the tolerance for overall transportation cost $\epsilon_c = 2.0$.
The resolution $r$ for the rasterization of the TIN model to generate heightmaps is dataset dependent due to the difference in point density. It is set to 0.20 m from AHN3, 0.15 m for DALES, and 0.25 m for Vaihingen.
The weight of the \textit{{roof preference}} energy term $\lambda_{r} = 0.04$ (while the weights for the data fitting and model complexity terms are set to $\lambda_{d}=0.34$ and $\lambda_{c}=0.62$, respectively).

\subsection{Comparisons}

\begin{figure}[H]

\begin{subfigure}[t]{0.24\linewidth}
\includegraphics[width=\linewidth]{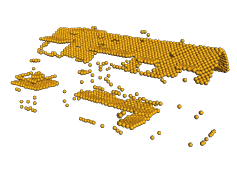}
\caption{\centering Input point cloud}
\end{subfigure}
\begin{subfigure}[t]{0.24\linewidth}
\includegraphics[width=\linewidth]{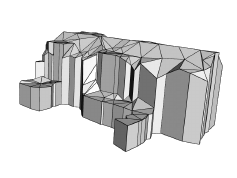}
\caption{\centering 2.5DC (296 faces)}
\end{subfigure}
\begin{subfigure}[t]{0.24\linewidth}
\includegraphics[width=\linewidth]{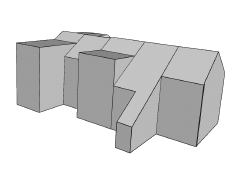}
\caption{\centering PolyFit (58 faces)}
\end{subfigure}
\begin{subfigure}[t]{0.24\linewidth}
\includegraphics[width=\linewidth]{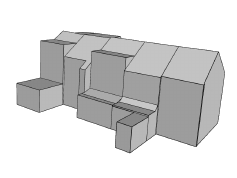}
\caption{\centering Ours (86 faces)}
\end{subfigure}
\caption{Comparison with 2.5D Dual Contouring {(2.5DC)}~\citep{zhou20102} 
and PolyFit~\citep{nan2017polyfit} on a single building from the AHN3 dataset~\cite{AHN3_2018}.}
\label{fig:comparison}
\end{figure}

We have compared our method with two successful open-source methods, i.e., 2.5D Dual Contouring (dedicated for urban buildings)~\citep{zhou20102} and PolyFit (for general piecewise-planar objects)~\citep{nan2017polyfit}, on the AHN3 ~\cite{AHN3_2018}, DALES~\cite{varney2020dales}, and Vaihingen~\cite{rottensteiner2012isprs} datasets.
{	The city block from the AHN3 dataset~\cite{AHN3_2018} is sparse and contains only 80,447 points for 160~buildings (i.e., on average 503 points per building).
The city region from DALES is denser and contains  214,601 points for 41 buildings  (i.e., on average  5234 points per building).
The city area from the Vaihingen dataset contains  69,254 points for 57 buildings (i.e., on average 1215~points per building).
The walls of all the point clouds are severely occluded.}
Figure~\ref{fig:comparison} shows the visual comparison of one of the buildings.
PolyFit assumes a complete set of input planar primitives, which is not the case for airborne LiDAR point clouds because the vertical walls are often missing.
For PolyFit to be effective, we added our inferred vertical planes to its initial set of planar primitives.
From the result, we can observe that both PolyFit and our method can generate compact building models, and the number of faces in the result is an order of magnitude less than that of the 2.5D Dual Contouring method.
It is worth noting that even with the additional planes, PolyFit still failed to reconstruct some walls and performed poorly in recovering geometric details.
In contrast, our method produces the most plausible 3D models. By inferring missing vertical planes, our method can recover inner walls, which further split the roof planes and bring in more geometric details into the final reconstruction.
{Table~\ref{tab:statistical_comparison} reports the statistics of the comparison, from which we can see that the reconstructed building models from our method have the highest accuracy.
In terms of running time, our method is slower than the other two, but it is still acceptable in practical applications (on average 4.9 s per building).}

\vspace{-6pt}
\begin{figure}[H]

\subfloat[\centering Result from BAG3D]{\includegraphics[width=1.0\linewidth]{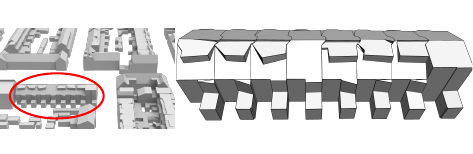}\label{subfig:bag3d}}
\vspace{0.5cm}
\subfloat[\centering Ours result]{\includegraphics[width=1.0\linewidth]{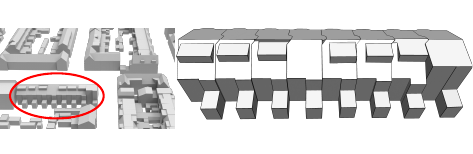}\label{subfig:ours}}
\caption{A visual comparison with {BAG3D}~\citep{BAG3D}. A building from 
Table~\ref{tab:statistical_comparison_BAG3D} {({b})} is shown.}
\label{fig:comparison_3DBAG}
\end{figure}

\vspace{-9pt}
\begin{table}[H]
\caption{{Statistics on the comparison of 2.5D Dual {Contouring}~\citep{zhou20102}, 
PolyFit~\citep{nan2017polyfit}, and our method on the reconstruction  from the AHN3 ~\cite{AHN3_2018}, DALES~\cite{varney2020dales}, and Vaihingen~\cite{rottensteiner2012isprs} datasets. Total face numbers, running times, and average errors are reported.}
}
\label{tab:statistical_comparison}
\setlength{\cellWidtha}{\textwidth/5-2\tabcolsep-0in}
\setlength{\cellWidthb}{\textwidth/5-2\tabcolsep-0in}
\setlength{\cellWidthc}{\textwidth/5-2\tabcolsep-0in}
\setlength{\cellWidthd}{\textwidth/5-2\tabcolsep-0in}
\setlength{\cellWidthe}{\textwidth/5-2\tabcolsep-0in}
\scalebox{1}[1]{\begin{tabularx}{\textwidth}{>{\centering\arraybackslash}m{\cellWidtha}>{\centering\arraybackslash}m{\cellWidthb}>{\centering\arraybackslash}m{\cellWidthc}>{\centering\arraybackslash}m{\cellWidthd}>{\centering\arraybackslash}m{\cellWidthe}}
\toprule
\textbf{Dataset} & \textbf{Method}
&   \textbf{ \#Faces}\ & \textbf{RMSE (m)} & \textbf{Time (s)} \\
\midrule
\multirow{4}{*}{AHN3}
&2.5D DC~\citep{zhou20102}   & 12,781                                                           & 0.213            & 13      \\  \cmidrule{2-5}
&PolyFit~\citep{nan2017polyfit} & 1848                                                            & 0.242            & 160      \\ \cmidrule{2-5}
&Ours    & 2453                                                           & 0.128            & 380      \\ \midrule
\multirow{4}{*}{DALES}
&2.5D DC~\citep{zhou20102}   &       2297                                                     &         0.204  &    10  \\  \cmidrule{2-5}
&PolyFit~\citep{nan2017polyfit} &   444                                                          &    0.287       &    230  \\ \cmidrule{2-5}
&Ours    &        583                                                    &       0.184    &   670   \\ \midrule
\multirow{4}{*}{Vaihingen}
&2.5D DC~\citep{zhou20102}   & 2695                                                           & 0.168            & 6      \\  \cmidrule{2-5}
&PolyFit~\citep{nan2017polyfit} & 647                                                            & 0.275            & 102      \\ \cmidrule{2-5}
&Ours    & 798                                                           & 0.157            & 212      \\ \bottomrule
\end{tabularx}}

\end{table}
We also performed an extensive quantitative comparison with the 3D building models from the BAG3D~\citep{BAG3D}, which is a public 3D city platform that provides 3D models of urban buildings at the LoD2 level. For this comparison, we picked four different regions consisting of 1113 buildings in total from the BAG3D.
In Figure~\ref{fig:comparison_3DBAG}, we demonstrate a visual comparison, from which we can see that our models demonstrate more regularity.
The quantitive result is reported in Table~\ref{tab:statistical_comparison_BAG3D}, from which we can see that our results have \mbox{higher accuracy}.

\vspace{-9pt}
\begin{table}[H]
\caption{Quantitative comparison with the {BAG3D}~\cite{BAG3D} on four urban scenes (a)--(d). Both BAG3D and our method used the point clouds from the AHN3 dataset~\cite{AHN3_2018} as input. The bold font indicates smaller RMSE values.
}
\label{tab:statistical_comparison_BAG3D}
\setlength{\cellWidtha}{\textwidth/5-2\tabcolsep-0in}
\setlength{\cellWidthb}{\textwidth/5-2\tabcolsep-0in}
\setlength{\cellWidthc}{\textwidth/5-2\tabcolsep-0in}
\setlength{\cellWidthd}{\textwidth/5-2\tabcolsep-0in}
\setlength{\cellWidthe}{\textwidth/5-2\tabcolsep-0in}
\scalebox{1}[1]{\begin{tabularx}{\textwidth}{>{\centering\arraybackslash}m{\cellWidtha}>{\centering\arraybackslash}m{\cellWidthb}>{\centering\arraybackslash}m{\cellWidthc}>{\centering\arraybackslash}m{\cellWidthd}>{\centering\arraybackslash}m{\cellWidthe}}
\toprule
\textbf{Region} &
\textbf{\#Points} &
\textbf{\#Building} &
\textbf{RMSE (m)\linebreak BAG3D} &
\textbf{RMSE (m)\linebreak Ours} \\ \midrule
(a) & 1,694,247 & 198 & 0.088 & \textbf{{0.079}} \\ \midrule 
(b) & 329,593  & 387 & 0.139 & \textbf{0.138} \\ \midrule
(c) & 224,970  & 368 & 0.140 & \textbf{0.132} \\ \midrule
(d) & 80,447   & 160 & 0.146 & \textbf{0.128} \\ \bottomrule
\end{tabularx}}

\end{table}

\subsection{\textit{With} vs. \textit{Without} Footprint}

\begin{figure}[H]

\includegraphics[width=0.9\linewidth]{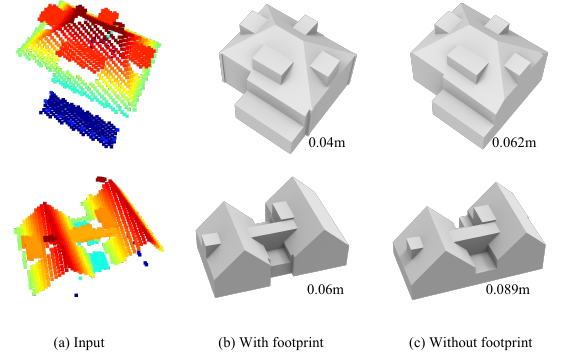}
\caption{{Comparison} 
between the reconstruction \emph{{with}} {(\textbf{b})}
and \emph{{without}} (\textbf{c}) footprint data on two buildings (\textbf{a}) from the AHN3 {dataset}~\cite{AHN3_2018}. The number below each model denotes the root mean square error (RMSE). Using the inferred vertical planes slightly increases reconstruction errors.}
\label{fig:comparison_without_fooprint}
\end{figure}

Our method can infer the vertical planes of a building from its roof points, and then the outer walls are completed using the vertical planes. It also has the option to directly use given footprint data for reconstruction. With a given footprint, vertically planes are firstly obtained by extruding the footprint polygons. Then these planes and those extracted from the point clouds are intersected to hypothesize the model faces, followed by the optimization step to obtain the final reconstruction.
Figure~\ref{fig:comparison_without_fooprint} shows such a comparison on two buildings.

\subsection{Reconstruction Using Point Clouds with Vertical Planes}
\begin{figure}[H]

\includegraphics[width=0.9\linewidth]{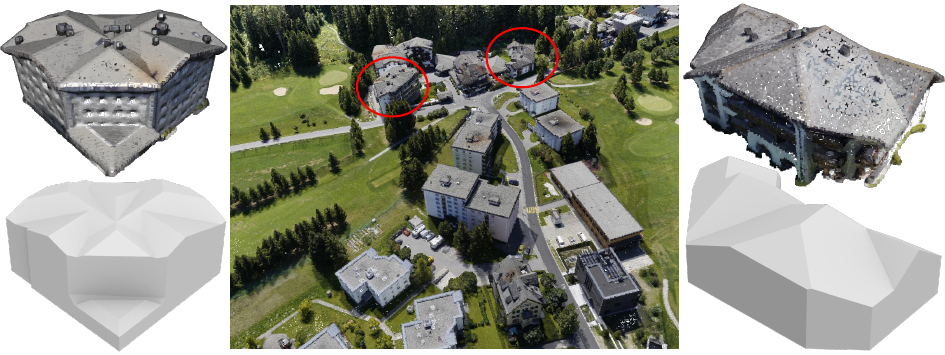}
\caption{Reconstruction from aerial point clouds. In these point clouds, the vertical walls can be extracted from the point clouds and directly used in reconstruction, and thus the vertical plane inference step was skipped. The dataset is obtained {from}~\citet{can2021semantic}.}
\label{fig:aerial_pointcloud}
\end{figure}

The methodology presented in our paper only focuses on airborne LiDAR point clouds, in which vertical walls of buildings are typically missing.
In practice, our method can be easily adapted to work with other types of point clouds that contain points of vertical walls, e.g., point clouds reconstructed from drone images. For such point clouds, our method can still be effective by replacing the inferred vertical planes with those directly detected from the point clouds. Figure~\ref{fig:aerial_pointcloud} shows two such examples.

\vspace{-9pt}

\subsection{Limitations}
\label{sec:limitation}

Our method can infer the missing vertical planes of buildings, from which the outer vertical planes serve as outer walls in the reconstruction. Since the vertical planes are inferred from the 3D points of rooftops, the walls in the final models may not perfectly align with the ground-truth footprints (see the figure {below}). Thus, we recommend the use of high-quality footprint data whenever it is available. 
Besides, our method extends the hypothesis-and-selection-based surface reconstruction framework of PolyFit~\cite{nan2017polyfit} by introducing new energy terms and hard constraints. It naturally inherits the limitation of PolyFit, i.e., it may encounter computation bottlenecks for buildings with complex structures (e.g., buildings with more than 100 planar regions). An example has already been shown in Figure~\ref{fig:individual_buildings_results} (12).
\vspace{-5pt}
\begin{figure}[H]
\includegraphics[width=0.21\textwidth]{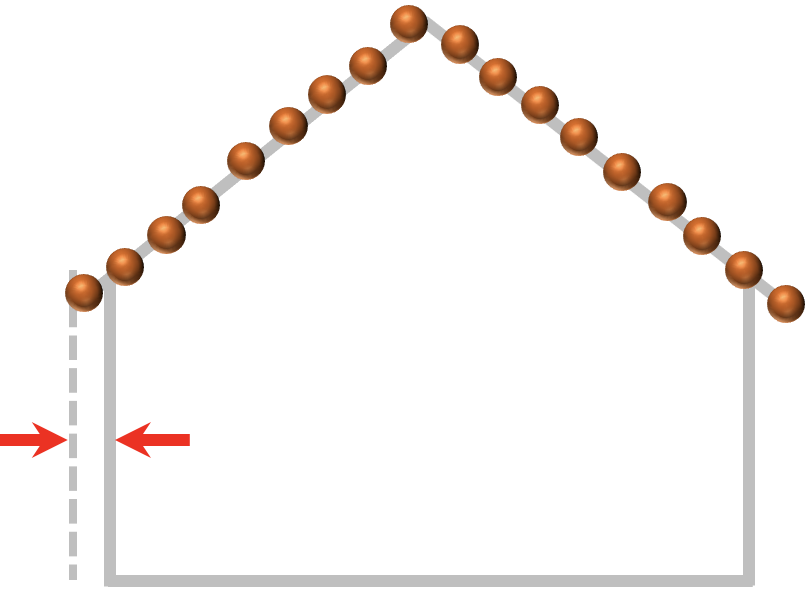}
\end{figure}

\section{Conclusions {and Future Work}}
\label{sec:conclusion}

We have presented a fully automatic approach for large-scale 3D reconstruction of urban buildings from airborne LiDAR point clouds.
We propose to infer the vertical planes of buildings that are commonly missing from airborne LiDAR point clouds. The inferred vertical planes play two different roles during the reconstruction. The outer vertical planes directly become part of the exterior walls of the building, and the inner vertical planes enrich building details by splitting the roof planes at proper locations and forming the necessary inner walls in final models. Our method can also incorporate given building footprints for reconstruction. In case footprints are used, they are extruded to serve the exterior walls of the models, and the inferred inner planes enrich building details.
Extensive experiments on different datasets have demonstrated that inferring vertical planes is an effective strategy for building reconstruction from airborne LiDAR point clouds, and the proposed \textit{{roof preference}} energy term and the novel hard constraints ensure topologically correct and accurate reconstruction.

{Our current framework uses only planar primitives and it is sufficient for reconstructing most urban buildings.
In the real world, there still exist buildings with curved surfaces, which our current implementation could not handle. However, our hypothesize-and-selection strategy is general and can be extended to process different types of primitives. As a future work direction, our method can be extended to incorporate other geometric primitives, such as spheres, cylinders, or even parametric surfaces. With such an extension, buildings with curved surfaces can also be reconstructed.}

\vspace{6pt}



\authorcontributions{J.H. performed the study and implemented the algorithms. R.P. and J.S. provided
constructive comments and suggestions. L.N. proposed this topic, provided daily supervision, and wrote the paper together with J.H. All authors have read and agreed to the published version of the manuscript.}

\funding{{Jin Huang is financially supported by the China Scholarship Council}.}

\dataavailability{Our code and data are available at~\url{https://github.com/yidahuang/City3D}, accessed on 23 March 2022.} 

\acknowledgments{We thank Zexin Yang, Zhaiyu Chen, and Noortje van der Horst for proofreading the paper. }

\conflictsofinterest{The authors declare no conflict of interest.}


\abbreviations{Abbreviations}{
The following abbreviations are used in this manuscript:\\

\noindent
\begin{tabular}{@{}ll}
LiDAR & Light Detection and Ranging\\
TIN & Triangular Irregular Network\\
RMSE & Root Mean Square Error \\

\end{tabular}}

\appendixtitles{yes} 
\appendixstart
\appendix

\section{{The Complete Formulation}
}
Our reconstruction is obtained by finding the optimal subset of the hypothesized faces.
We formulate this as an optimization problem, with an objective function consisting of three energy terms: \textit{{data fitting}}, \textit{{model complexity}}, and \textit{{roof preference}}. The first two terms are the same as in~\cite{nan2017polyfit}. In the following, we briefly introduce all these terms and provide the final complete formulation.

\begin{itemize}
\item \textbf{{Data fitting}}. It is defined to measure how well the final model (i.e., the assembly of the chosen faces) fits to the input point cloud,
\begin{equation}\label{term1}
E_{d}=1-\frac{1}{\left|P\right|}\sum_{i=1}^{|F|}x_{i}\cdot support (f_{i}),
\end{equation}
where $ |P| $ is the number of points in the point cloud.
$ support (f_{i}) $ measures the number of points that are $\epsilon$-close to a face $ f_i \in F $, and $x_i \in \{0, 1\}$ denotes the binary status of the face $f_i$ (1 for \textit{{selected}} and 0 otherwise). $|F|$ denotes the total number of \mbox{hypothesized faces}.

\item \textbf{{Model complexity}}. To avoid defects introduced by noise and outliers, this term is introduced to encourage large planar structures,
\begin{equation}\label{term2}
E_{c}= \frac{1}{\left|E\right|}\sum_{i=1}^{\left|E\right|} corner (e_{i}),
\end{equation}
where $ |E| $ denotes the total number of pairwise intersections in the hypothesized face set.
$ corner(e_i) $ is an indicator function denoting if choosing two faces connected by an edge $e_i$ results in a sharp edge in the final model (1 for \textit{{sharp}} and 0 otherwise).

\item \textbf{{Roof preference}}.
We have observed in rare cases that a building in aerial point clouds may demonstrate more than one layer of roofs, e.g., semi-transparent or overhung roofs. In such a case, we assume a higher roof face is preferable to the ones underneath. We formulate this preference as an additional \textit{{roof preference}} energy term,
\begin{equation}\label{term3}
E_{r}=\frac{1}{\left| F\right| }\sum_{i=1}^{|F|}x_{i} \cdot \frac{z_{max}-z_{i}}{z_{max}-z_{min}}
\end{equation}
where $z_{i} $ denotes the $Z$ coordinate of the centroid of a face $f_{i}$. $z_{max}$ and $z_{min}$ are, respectively, the highest and lowest $Z$ coordinates of the building points.
\end{itemize}

With all the constraints, the complete optimization problem is written as
\begin{equation}	\label{eq:optimization}
\begin{aligned}
&\min_{X} \lambda_{d}E_{d}+\lambda_{c} E_{c}+\lambda_{r} E_{r}  \\
&\textrm{s.t.} \quad  \left\{
\begin{array}{lr}
\sum\limits_{k \in V(f_{i})} x_{k}=1 ,  (1 \leq i \leq |F|)\\
\sum \limits_{j\in N(e_i) } x_{j}=0 \quad or \quad 2 , \quad (1 \leq j \leq |E|)\\
x_{l}=1,\\
x_{i}\in \{0,1\}, \quad \forall i\in N  \\
\end{array}
\right.\\
\end{aligned}
\end{equation}
where the first constraint is call \textit{{single roof}}, which ensures that the reconstructed building model has a single layer of roofs. The second constraint enforces that in the final model an edge is associated with two adjacent faces, ensuring the final model to be watertight and manifold. The third constraint is call \textit{{face prior}}, which ensures that, for the faces derived from the same planar segment, the one with the highest confidence value is selected as \mbox{a prior}.

By solving the above optimization problem, the set of selected faces $\{f_i | x_i = 1\}$ forms the final surface model of a building.

\begin{adjustwidth}{-\extralength}{0cm}

\reftitle{{References}}
\end{adjustwidth}

\begin{thebibliography}{999}

\bibitem[Yao {et~al.}(2018)Yao, Nagel, Kunde, Hudra, Willkomm, Donaubauer,
Adolphi, and Kolbe]{yao20183dcitydb}
Yao, Z.; Nagel, C.; Kunde, F.; Hudra, G.; Willkomm, P.; Donaubauer, A.;
Adolphi, T.; Kolbe, T.H.
\newblock 3DCityDB---A 3D geodatabase solution for the management, analysis, and
visualization of semantic 3D city models based on CityGML.
\newblock {\em Open Geospat. Data Softw. Stand.} {\bf 2018}, {\em
3},~1--26. [\href{http://doi.org/10.1186/s40965-018-0046-7}{CrossRef}]

\bibitem[Zhivov {et~al.}(2017)Zhivov, Case, Jank, Eicker, and
Booth]{zhivov2017planning}
Zhivov, A.M.; Case, M.P.; Jank, R.; Eicker, U.; Booth, S.
\newblock Planning tools to simulate and optimize neighborhood energy systems.
In {\em Green Defense Technology}; { Springer:  Dordrecht, The Netherlands, 2017;} pp. 137--163. 

\bibitem[Stoter {et~al.}(2020)Stoter, Peters, Commandeur, Dukai, Kumar, and
Ledoux]{stoter2020automated}
Stoter, J.; Peters, R.; Commandeur, T.; Dukai, B.; Kumar, K.; Ledoux, H.
\newblock Automated reconstruction of 3D input data for noise simulation.
\newblock {\em Comput. Environ. Urban Syst.} {\bf 2020}, {\em
80},~101424. [\href{http://dx.doi.org/10.1016/j.compenvurbsys.2019.101424}{CrossRef}]

\bibitem[Widl {et~al.}(2021)Widl, Agugiaro, and
Peters-Anders]{widl2021linking}
Widl, E.; Agugiaro, G.; Peters-Anders, J.
\newblock Linking Semantic 3D City Models with Domain-Specific Simulation Tools
for the Planning and Validation of Energy Applications at District Level.
\newblock {\em Sustainability} {\bf 2021}, {\em 13},~8782. [\href{http://dx.doi.org/10.3390/su13168782}{CrossRef}]

\bibitem[Cappelle {et~al.}(2012)Cappelle, El~Najjar, Charpillet, and
Pomorski]{cappelle2012virtual}
Cappelle, C.; El~Najjar, M.E.; Charpillet, F.; Pomorski, D.
\newblock Virtual 3D city model for navigation in urban areas.
\newblock {\em J. Intell. Robot. Syst.} {\bf 2012}, {\em
66},~377--399. [\href{http://dx.doi.org/10.1007/s10846-011-9594-0}{CrossRef}]

\bibitem[Kargas {et~al.}(2019)Kargas, Loumos, and Varoutas]{kargas2019using}
Kargas, A.; Loumos, G.; Varoutas, D.
\newblock Using different ways of 3D reconstruction of historical cities for
gaming purposes: The case study of Nafplio.
\newblock {\em Heritage} {\bf 2019}, {\em 2},~1799--1811. [\href{http://dx.doi.org/10.3390/heritage2030110}{CrossRef}]

\bibitem[Nan {et~al.}(2010)Nan, Sharf, Zhang, Cohen-Or, and
Chen]{nan2010smartboxes}
Nan, L.; Sharf, A.; Zhang, H.; Cohen-Or, D.; Chen, B.
\newblock Smartboxes for interactive urban reconstruction. In {\em ACM Siggraph
2010 Papers};  {ACM: New Yrok, NY, USA, 2010;} 
pp. 1--10.

\bibitem[Nan {et~al.}(2015)Nan, Jiang, Ghanem, and Wonka]{nan2015template}
Nan, L.; Jiang, C.; Ghanem, B.; Wonka, P.
\newblock Template assembly for detailed urban reconstruction.
\newblock  In \emph{Computer Graphics Forum}; {Wiley Online Library:  Zurich, Switzerland, 2015;} 
Volume~34, pp.
217--228.

\bibitem[Zhou(2012)]{zhou20123d}
Zhou, Q.Y.
\newblock {\em 3D Urban Modeling from City-Scale Aerial LiDAR Data}; {University
of Southern California:  Los Angeles, CA, USA, 2012.}

\bibitem[Haala {et~al.}(2015)Haala, Rothermel, and
Cavegn]{haala2015extracting}
Haala, N.; Rothermel, M.; Cavegn, S.
\newblock Extracting 3D urban models from oblique aerial images.
\newblock  In Proceedings of the 2015 Joint Urban Remote Sensing Event (JURSE),  Lausanne, Switzerland, 30 March--1 April 2015; pp.
1--4.

\bibitem[Verdie {et~al.}(2015)Verdie, Lafarge, and Alliez]{verdie2015lod}
Verdie, Y.; Lafarge, F.; Alliez, P.
\newblock LOD generation for urban scenes.
\newblock {\em ACM Trans. Graph.} {\bf 2015}, {\em 34},~30. [\href{http://dx.doi.org/10.1145/2732527}{CrossRef}]

\bibitem[Li {et~al.}(2016)Li, Nan, Smith, and Wonka]{li2016reconstructing}
Li, M.; Nan, L.; Smith, N.; Wonka, P.
\newblock Reconstructing building mass models from UAV images.
\newblock {\em Comput. Graph.} {\bf 2016}, {\em 54},~84--93. [\href{http://dx.doi.org/10.1016/j.cag.2015.07.004}{CrossRef}]

\bibitem[Buyukdemircioglu {et~al.}(2018)Buyukdemircioglu, Kocaman, and
Isikdag]{buyukdemircioglu2018semi}
Buyukdemircioglu, M.; Kocaman, S.; Isikdag, U.
\newblock Semi-automatic 3D city model generation from large-format aerial
images.
\newblock {\em ISPRS Int. J.-Geo-Inf.} {\bf 2018}, {\em
7},~339. [\href{http://dx.doi.org/10.3390/ijgi7090339}{CrossRef}]

\bibitem[Bauchet and Lafarge(2019)]{bauchet2019}
Bauchet, J.P.; Lafarge, F.
\newblock {City Reconstruction from Airborne Lidar: A Computational Geometry
Approach}.
\newblock In Proceedings of the  {3D GeoInfo 2019---14thConference 3D GeoInfo},  {Singapore, 26--27 September 2019.} 


\bibitem[Li {et~al.}(2019)Li, Rottensteiner, and Heipke]{li2019modelling}
Li, M.; Rottensteiner, F.; Heipke, C.
\newblock Modelling of buildings from aerial LiDAR point clouds using TINs and
label maps.
\newblock {\em ISPRS J. Photogramm. Remote Sens.} {\bf 2019},
{\em 154},~127--138. [\href{http://dx.doi.org/10.1016/j.isprsjprs.2019.06.003}{CrossRef}]

\bibitem[Ledoux {et~al.}(2021)Ledoux, Biljecki, Dukai, Kumar, Peters,
Stoter, and Commandeur]{ledoux20213dfier}
Ledoux, H.; Biljecki, F.; Dukai, B.; Kumar, K.; Peters, R.; Stoter, J.;
Commandeur, T.
\newblock 3dfier: Automatic reconstruction of 3D city models.
\newblock {\em J. Open Source Softw.} {\bf 2021}, {\em 6},~2866. [\href{http://dx.doi.org/10.21105/joss.02866}{CrossRef}]

\bibitem[Zhou {et~al.}(2020)Zhou, Yi, Liu, Huang, and Huang]{zhou2020survey}
Zhou, X.; Yi, Z.; Liu, Y.; Huang, K.; Huang, H.
\newblock Survey on path and view planning for UAVs.
\newblock {\em Virtual Real. Intell. Hardw.} {\bf 2020}, {\em
2},~56--69. [\href{http://dx.doi.org/10.1016/j.vrih.2019.12.004}{CrossRef}]

\bibitem[Qi {et~al.}(2017)Qi, Su, Mo, and Guibas]{qi2017pointnet}
Qi, C.R.; Su, H.; Mo, K.; Guibas, L.J.
\newblock Pointnet: Deep learning on point sets for 3d classification and
segmentation.
\newblock  In Proceedings of the IEEE Conference on Computer Vision and Pattern
Recognition,  {Honolulu, HI, USA,  21--26 July 2017;} 
pp. 652--660.

\bibitem[Thomas {et~al.}(2019)Thomas, Qi, Deschaud, Marcotegui, Goulette,
and Guibas]{thomas2019kpconv}
Thomas, H.; Qi, C.R.; Deschaud, J.E.; Marcotegui, B.; Goulette, F.; Guibas,
L.J.
\newblock Kpconv: Flexible and deformable convolution for point clouds.
\newblock  In Proceedings of the IEEE/CVF International Conference on Computer
Vision, { Seoul, Korea, 27 October--\mbox{2 November 2019};} 
pp. 6411--6420.

\bibitem[Nan and Wonka(2017)]{nan2017polyfit}
Nan, L.; Wonka, P.
\newblock PolyFit: Polygonal Surface Reconstruction from Point Clouds.
\newblock   In Proceedings of the IEEE International Conference on Computer Vision,  {Venice, Italy, 22--29 October, 2017.}

\bibitem[{AHN3}(2018)]{AHN3_2018}
{AHN3}.
\newblock {Actueel Hoogtebestand Nederland (AHN)}. 2018.
\newblock Available online: \url{https://www.pdok.nl/ nl/ahn3-downloads} {(\mbox{accessed on 13 November 2021}).} 


\bibitem[Fischler and Bolles(1981)]{fischler1981random}
Fischler, M.A.; Bolles, R.C.
\newblock Random sample consensus: A paradigm for model fitting with
applications to image analysis and automated cartography.
\newblock {\em Commun.  ACM} {\bf 1981}, {\em 24},~381--395. [\href{http://dx.doi.org/10.1145/358669.358692}{CrossRef}]

\bibitem[Schnabel {et~al.}(2007)Schnabel, Wahl, and
Klein]{schnabel2007efficient}
Schnabel, R.; Wahl, R.; Klein, R.
\newblock Efficient RANSAC for point-cloud shape detection.
\newblock In \emph{Computer Graphics Forum}; {Wiley Online Library:  Oxford, UK, 2007}; Volume~26, pp.
214--226.


\bibitem[Zuliani {et~al.}(2005)Zuliani, Kenney, and
Manjunath]{zuliani2005multiransac}
Zuliani, M.; Kenney, C.S.; Manjunath, B.
\newblock The multiransac algorithm and its application to detect planar
homographies.
\newblock  \mbox{In Proceedings of the} IEEE International Conference on Image Processing 2005,  Genova, Italy, 14 September 2005;
Volume~3, \mbox{p. {III-153.}} 


\bibitem[Rabbani {et~al.}(2006)Rabbani, Van Den~Heuvel, and
Vosselmann]{rabbani2006segmentation}
Rabbani, T.; Van Den~Heuvel, F.; Vosselmann, G.
\newblock Segmentation of point clouds using smoothness constraint.
\newblock {\em Int. Arch. Photogramm. Remote Sens. Spat. Inf. Sci.} {\bf 2006}, {\em 36},~248--253.

\bibitem[Sun and Salvaggio(2013)]{sun2013aerial}
Sun, S.; Salvaggio, C.
\newblock Aerial 3D building detection and modeling from airborne LiDAR point
clouds.
\newblock {\em IEEE J. Sel. Top. Appl. Earth Obs. Remote Sens.} {\bf 2013}, {\em 6},~1440--1449. [\href{http://dx.doi.org/10.1109/JSTARS.2013.2251457}{CrossRef}]

\bibitem[Chen {et~al.}(2017)Chen, Wang, and
Peethambaran]{chen2017topologically}
Chen, D.; Wang, R.; Peethambaran, J.
\newblock Topologically aware building rooftop reconstruction from airborne
laser scanning point clouds.
\newblock {\em IEEE Trans. Geosci. Remote Sens.} {\bf 2017},
{\em 55},~7032--7052. [\href{http://dx.doi.org/10.1109/TGRS.2017.2738439}{CrossRef}]

\bibitem[Meng {et~al.}(2009)Meng, Wang, and Currit]{meng2009morphology}
Meng, X.; Wang, L.; Currit, N.
\newblock Morphology-based building detection from airborne LIDAR data.
\newblock {\em Photogramm. Eng. Remote Sens.} {\bf 2009}, {\em
75},~437--442. [\href{http://dx.doi.org/10.14358/PERS.75.4.437}{CrossRef}]

\bibitem[Douglas and Peucker(1973)]{douglas1973algorithms}
Douglas, D.H.; Peucker, T.K.
\newblock Algorithms for the reduction of the number of points required to
represent a digitized line or its caricature.
\newblock {\em Cartogr. Int. J. Geogr. Inf. Geovis.} {\bf 1973}, {\em 10},~112--122. [\href{http://dx.doi.org/10.3138/FM57-6770-U75U-7727}{CrossRef}]

\bibitem[Zhang {et~al.}(2006)Zhang, Yan, and Chen]{zhang2006automatic}
Zhang, K.; Yan, J.; Chen, S.C.
\newblock Automatic construction of building footprints from airborne LIDAR
data.
\newblock {\em IEEE Trans. Geosci. Remote Sens.} {\bf 2006},
{\em 44},~2523--2533. [\href{http://dx.doi.org/10.1109/TGRS.2006.874137}{CrossRef}]

\bibitem[Xiong {et~al.}(2016)Xiong, Elberink, and
Vosselman]{xiong2016footprint}
Xiong, B.; Elberink, S.O.; Vosselman, G.
\newblock Footprint map partitioning using airborne laser scanning data.
\newblock {\em ISPRS Ann. Photogramm. Remote Sens. Spat. Inf. Sci.} {\bf 2016}, {\em 3}, 241--247. [\href{http://dx.doi.org/10.5194/isprs-annals-III-3-241-2016}{CrossRef}]

\bibitem[Zhou and Neumann(2008)]{zhou2008fast}
Zhou, Q.Y.; Neumann, U.
\newblock Fast and extensible building modeling from airborne LiDAR data.
\newblock  In Proceedings of the 16th ACM SIGSPATIAL International Conference on
Advances in Geographic Information Systems,   Irvine, CA, USA, 5--7 	November  2008; pp. 1--8.

\bibitem[Dorninger and Pfeifer(2008)]{dorninger2008comprehensive}
Dorninger, P.; Pfeifer, N.
\newblock A comprehensive automated 3D approach for building extraction,
reconstruction, and regularization from airborne laser scanning point clouds.
\newblock {\em Sensors} {\bf 2008}, {\em 8},~7323--7343. [\href{http://dx.doi.org/10.3390/s8117323}{CrossRef}]

\bibitem[Lafarge and Mallet(2012)]{lafarge2012creating}
Lafarge, F.; Mallet, C.
\newblock Creating large-scale city models from 3D-point clouds: A robust
approach with hybrid representation.
\newblock {\em Int. J. Comput. Vis.} {\bf 2012}, {\em
99},~69--85. [\href{http://dx.doi.org/10.1007/s11263-012-0517-8}{CrossRef}]

\bibitem[Xiao {et~al.}(2015)Xiao, Wang, Li, Zhang, Xi, Wang, and
Dong]{xiao2015building}
Xiao, Y.; Wang, C.; Li, J.; Zhang, W.; Xi, X.; Wang, C.; Dong, P.
\newblock Building segmentation and modeling from airborne LiDAR data.
\newblock {\em Int. J. Digit. Earth} {\bf 2015}, {\em
8},~694--709. [\href{http://dx.doi.org/10.1080/17538947.2014.914252}{CrossRef}]

\bibitem[Yi {et~al.}(2017)Yi, Zhang, Wu, Xu, Remil, Wei, and
Wang]{yi2017urban}
Yi, C.; Zhang, Y.; Wu, Q.; Xu, Y.; Remil, O.; Wei, M.; Wang, J.
\newblock Urban building reconstruction from raw LiDAR point data.
\newblock {\em Comput.-Aided Des.} {\bf 2017}, {\em 93},~1--14. [\href{http://dx.doi.org/10.1016/j.cad.2017.07.005}{CrossRef}]

\bibitem[Zhou and Neumann(2010)]{zhou20102}
Zhou, Q.Y.; Neumann, U.
\newblock 2.5 d dual contouring: A robust approach to creating building models
from aerial lidar point clouds.
\newblock  In \emph{European Conference on Computer Vision}; {Springer: Cham, Switzerland, 2010}; pp.
115--128.

\bibitem[Zhou and Neumann(2011)]{zhou20112}
Zhou, Q.Y.; Neumann, U.
\newblock 2.5 D building modeling with topology control.
\newblock  In Proceedings of the CVPR 2011,  Colorado Springs, CO, USA, 20--25 June 2011; pp. 2489--2496.

\bibitem[Chauve {et~al.}(2010)Chauve, Labatut, and Pons]{chauve2010robust}
Chauve, A.L.; Labatut, P.; Pons, J.P.
\newblock Robust piecewise-planar 3D reconstruction and completion from
large-scale unstructured point data.
\newblock  In Proceedings of the 2010 IEEE Computer Society Conference on Computer Vision and Pattern
Recognition, San Francisco, CA, USA, 13--18 June  2010; pp. 1261--1268.

\bibitem[Lafarge {et~al.}(2008)Lafarge, Descombes, Zerubia, and
Pierrot-Deseilligny]{lafarge2008structural}
Lafarge, F.; Descombes, X.; Zerubia, J.; Pierrot-Deseilligny, M.
\newblock Structural approach for building reconstruction from a single DSM.
\newblock {\em IEEE Trans. Pattern Anal. Mach. Intell.}
{\bf 2008}, {\em 32},~135--147. [\href{http://dx.doi.org/10.1109/TPAMI.2008.281}{CrossRef}]

\bibitem[Xiong {et~al.}(2014)Xiong, Elberink, and Vosselman]{xiong2014graph}
Xiong, B.; Elberink, S.O.; Vosselman, G.
\newblock A graph edit dictionary for correcting errors in roof topology graphs
reconstructed from point clouds.
\newblock {\em ISPRS J. Photogramm. Remote Sens.} {\bf 2014},
{\em 93},~227--242. [\href{http://dx.doi.org/10.1016/j.isprsjprs.2014.01.007}{CrossRef}]

\bibitem[Li {et~al.}(2016)Li, Wonka, and Nan]{li2016boxfitting}
Li, M.; Wonka, P.; Nan, L.
\newblock Manhattan-world Urban Reconstruction from Point Clouds.
\newblock  In \emph{European Conference on Computer Vision};  Springer: Cham, Switzerland, 2016.

\bibitem[Bauchet and Lafarge(2020)]{bauchet2020kinetic}
Bauchet, J.P.; Lafarge, F.
\newblock Kinetic shape reconstruction.
\newblock {\em ACM Trans. Graph. (TOG)} {\bf 2020}, {\em 39},~1--14. [\href{http://dx.doi.org/10.1145/3376918}{CrossRef}]

\bibitem[Fang and Lafarge(2020)]{fang2020connect}
Fang, H.; Lafarge, F.
\newblock Connect-and-Slice: An hybrid approach for reconstructing 3D objects.
\newblock In Proceedings of the IEEE/CVF Conference on Computer Vision and
Pattern Recognition,  {Seattle, WA, USA, 14--19 June 2020;} 
pp. 13490--13498.

\bibitem[Huang {et~al.}(2013)Huang, Brenner, and
Sester]{huang2013generative}
Huang, H.; Brenner, C.; Sester, M.
\newblock A generative statistical approach to automatic 3D building roof
reconstruction from laser scanning data.
\newblock {\em ISPRS J. Photogramm. Remote Sens.} {\bf 2013},
{\em 79},~29--43. [\href{http://dx.doi.org/10.1016/j.isprsjprs.2013.02.004}{CrossRef}]

\bibitem[Canny(1986)]{canny1986computational}
Canny, J.
\newblock A computational approach to edge detection.
\newblock {\em IEEE Trans. Pattern Anal. Mach. Intell.}
{\bf 1986}, \emph{PAMI-8}, 679--698. [\href{http://dx.doi.org/10.1109/TPAMI.1986.4767851}{CrossRef}]

\bibitem[De~Goes {et~al.}(2011)De~Goes, Cohen-Steiner, Alliez, and
Desbrun]{de2011optimal}
De~Goes, F.; Cohen-Steiner, D.; Alliez, P.; Desbrun, M.
\newblock An optimal transport approach to robust reconstruction and
simplification of 2D shapes.
\newblock  In \emph{Computer Graphics Forum}; {Wiley Online Library:  Oxford, UK, 2011}; Volume~30, pp.
1593--1602.

\bibitem[Li and Wu(2021)]{li2021relation}
Li, Y.; Wu, B.
\newblock Relation-Constrained 3D Reconstruction of Buildings in Metropolitan
Areas from Photogrammetric Point Clouds.
\newblock {\em Remote Sens.} {\bf 2021}, {\em 13},~129. [\href{http://dx.doi.org/10.3390/rs13010129}{CrossRef}]

\bibitem[Schubert {et~al.}(2017)Schubert, Sander, Ester, Kriegel, and
Xu]{schubert2017dbscan}
Schubert, E.; Sander, J.; Ester, M.; Kriegel, H.P.; Xu, X.
\newblock DBSCAN revisited, revisited: Why and how you should (still) use
DBSCAN.
\newblock {\em ACM Trans. Database Syst. (TODS)} {\bf 2017}, {\em
42},~1--21. [\href{http://dx.doi.org/10.1145/3068335}{CrossRef}]

\bibitem[{ CGAL library}(2020)]{cgal:eb-20a}
{ CGAL Library}.
\newblock {\em {CGAL} User and Reference Manual}, {5.0.2} ed.; {CGAL Editorial
Board:  Valbonne, French, 2020.}

\bibitem[{BAG}(2019)]{BAG_2018}
{BAG}.
\newblock {Basisregistratie Adressen en Gebouwen (BAG)}. 2019.
\newblock Available online: \url{https://bag.basisregistraties.overheid.nl/datamodel} {(accessed on 13 November 2021)}. 

\bibitem[Varney {et~al.}(2020)Varney, Asari, and Graehling]{varney2020dales}
Varney, N.; Asari, V.K.; Graehling, Q.
\newblock DALES: A large-scale aerial LiDAR data set for semantic segmentation.
\newblock  In Proceedings of the IEEE/CVF Conference on Computer Vision and
Pattern Recognition Workshops,  {Seattle, WA, USA, 14--19 June 2020}; pp.~186--187.

\bibitem[Rottensteiner {et~al.}(2012)Rottensteiner, Sohn, Jung, Gerke,
Baillard, Benitez, and Breitkopf]{rottensteiner2012isprs}
Rottensteiner, F.; Sohn, G.; Jung, J.; Gerke, M.; Baillard, C.; Benitez, S.;
Breitkopf, U.
\newblock The ISPRS benchmark on urban object classification and 3D building
reconstruction.
\newblock {\em ISPRS Ann. Photogramm. Remote Sens. Spat. Inf. Sci. I-3} {\bf 2012}, {\em 1},~293--298. [\href{http://dx.doi.org/10.5194/isprsannals-I-3-293-2012}{CrossRef}]

\bibitem[Kazhdan {et~al.}(2006)Kazhdan, Bolitho, and
Hoppe]{kazhdan2006poisson}
Kazhdan, M.; Bolitho, M.; Hoppe, H.
\newblock Poisson surface reconstruction.
\newblock  In Proceedings of the Fourth Eurographics Symposium on Geometry
Processing,  {Cagliari, Italy, 26--28 June  2006;} Volume~7.

\bibitem[BAG(2021)]{BAG3D}
{3D BAG} (v21.09.8). 2021.
\newblock Available online: \url{https://3dbag.nl/en/viewer} {(accessed on  13 November 2021).} 

\bibitem[Can {et~al.}(2021)Can, Mantegazza, Abbate, Chappuis, and
Giusti]{can2021semantic}
Can, G.; Mantegazza, D.; Abbate, G.; Chappuis, S.; Giusti, A.
\newblock Semantic segmentation on Swiss3DCities: A benchmark study on aerial
photogrammetric 3D pointcloud dataset.
\newblock {\em Pattern Recognit. Lett.} {\bf 2021}, {\em 150},~108--114. [\href{http://dx.doi.org/10.1016/j.patrec.2021.06.004}{CrossRef}]

\end{thebibliography}
\end{document}